\documentclass[pdflatex,sn-basic]{sn-jnl}
\usepackage{graphicx}%
\usepackage{multirow}%
\usepackage{amsmath,amssymb,amsfonts}%
\usepackage{amsthm}%
\usepackage{mathrsfs}%
\usepackage[title]{appendix}%
\usepackage{xcolor}%
\usepackage{textcomp}%
\usepackage{manyfoot}%
\usepackage{subcaption}%
\usepackage{footmisc}
\usepackage[linesnumbered,ruled,vlined]{algorithm2e}%
\usepackage{algorithmicx}%
\usepackage{algpseudocode}%
\usepackage{listings}%
\usepackage{float}
\usepackage{bbm}

\theoremstyle{thmstyleone}%
\theoremstyle{thmstyletwo}%

\theoremstyle{thmstylethree}%

\title[Article Title]{K-Models: a Flexible and Interpretable Method for Ordinal Clustering with Application to Antigen-Antibody Interaction Profiles}

\author*[1]{\fnm{Giulia} \sur{Patanè}}\email{giulia.patane@polimi.it}

\author[1]{\fnm{Alessandra} \sur{Menafoglio}}

\author[3]{\fnm{Alexander} \sur{Krauth}}

\author[3]{\fnm{Peter} \sur{Fechner}}

\author[1]{\fnm{Luca} \sur{Dede'}}

\author[2]{\fnm{Bianca Maria} \sur{Colosimo}}

\author[1]{\fnm{Federica} \sur{Nicolussi}}

\affil*[1]{\orgdiv{MOX, Department of Mathematics}, \orgname{Politecnico di Milano}, \orgaddress{\street{Piazza Leonardo da Vinci 32}, \city{Milan}, \postcode{20133}, \country{Italy}}}

\affil[2]{\orgdiv{Department of Mechanical Engineering}, \orgname{Politecnico di Milano}, \orgaddress{\street{Piazza Leonardo da Vinci 32}, \city{Milan}, \postcode{20133}, \country{Italy}}}

\affil[3]{\orgdiv{Institute of Physical and Theoretical Chemistry (IPTC)}, \orgname{Eberhard Karls Universität Tübingen}, \orgaddress{\street{Auf der Morgenstelle 18}, \city{Tübingen}, \postcode{72076}, \country{Germany}}}
\begin{document}
\abstract{Existing clustering methods for functional data often prioritize partitioning accuracy over interpretability, making it challenging to extract meaningful insights when the data-generating process follows a specific underlying structure and an ordinal relationship among clusters is suspected. This work introduces K-Models, a novel framework that integrates ordinal constraints and estimates key underlying elements of the random process generating the observed functional profiles, improving both interpretability and structure identification. The proposed method is evaluated through simulations and real-world applications. In particular, it is tested on Region of Interest (ROI) curves, which represent reaction profiles from a reflectometric sensor monitoring biomolecular interactions, such as antigen-antibody binding. These curves represent changes in reflected light intensity over time at multiple measurement spots with immobilized antigens during analyte exposure, capturing the binding dynamics of the system. The goal is to identify intrinsic signal patterns solely from the observed dynamics, making this dataset an ideal benchmark for assessing the added interpretability of the proposed approach. By incorporating structural assumptions into the clustering process, K-Models enhances interpretability while maintaining performance comparable to state-of-the-art techniques, providing a valuable tool for analyzing functional data with an underlying ordinal structure.}

 \maketitle

\section{Introduction}\label{sec1}
\label{sec:introduction}
Functional data has become increasingly widespread in recent years, with applications in fields such as engineering \cite{engineering}, medicine \cite{medical,neuroscience}, psychometry \cite{psychometry}, and many others. As a result, Functional Data Analysis (FDA, \cite{ramsaysilvermanbook, kolmogorov1961,horvath12}) has gained significant attention, earning a central role in statistical research. For functional data we often, but not always, intend a set of curves defined in an infinite-dimensional space, where each curve represents a random function, which in turn is seen as the realisation of an underlying random process. Within this framework, functional clustering has also emerged as topic of interest, leading to the development of new methods aimed to partitioning curves based on their similarities \cite{zhang2022reviewclusteringmethodsfunctional}. Formally, functional clustering refers to the unsupervised task of grouping a set of curves in distinct subsets, named clusters, in such a way that instances within a group are similar to each other while they are dissimilar to instances of other groups (see for instance \cite{Wu18102022}). Various methods have been developed, each leveraging different characteristics of functional data. Given this broad range of techniques available, several classification schemes have been proposed in order to have a systematic description, such as those by \cite{JaquePreda} or \cite{zhang2022reviewclusteringmethodsfunctional}. Among the main categories, we can identify:
\begin{itemize}
    \item \textit{Raw data} methods, which cluster functions based on their observed values on a discrete grid of their domain. These approaches directly compare the functional observations without any prior transformation, making them conceptually straightforward. A well-known example is K-Means \citep{Oti2021, Bock2007}, a widely used algorithm which can be easily adapted to various settings.
    \item \textit{Filtering} methods, which consist of two phases: the first phase involves approximating curves using a basis expansion, which leads to a finite-dimensional representation, and the second phase involves performing clustering on the resulting finite-dimensional objects. One such method, introduced by \cite{Wu18102022}, begins by applying functional Principal Component Analysis (fPCA), projecting the functions onto principal component curves to obtain principal scores, which are subsequently used for classical multivariate clustering.
    \item \textit{Distance-based} methods, which rely on functional distance metrics to quantify similarities between curves. In reality, this category is overlapping with the previous two, as the choice of distance metric determines whether the method aligns more closely with raw data approaches or filtering approaches. An example is the PAM (Partitioning Around Medoids, \cite{Kaufman1990, Schubert_2019}) algorithm, or even K-Means itself, since they both are based on dissimilarity metrics that work directly with the observed values of the curves.
\end{itemize}

While existing functional clustering methods are designed to achieve well-separated groups, they rarely focus on uncovering the mechanisms driving the observed curves. We encountered this limitation when analysing a biosensor dataset of time-resolved biosensor signals, recording the interaction of an antibody with sensor spots functionalised with different antigen concentrations. The curves share a common pattern -- an initial stable phase followed by a rapid decline -- but differ in intensity. This variation is driven by differences in the immobilised antigen concentrations, since a single antibody concentration is applied uniformly across all antigens within each assay. The observed curves therefore reflect an ordinal latent structure, in which cluster membership indexes the magnitude of a common underlying effect rather than qualitatively distinct processes.

Traditional clustering methods, although capable of producing meaningful partitions, do not explicitly account for this ordinal nature. They divide the data into distinct groups but fail to estimate the latent components influencing curve generation. This challenge led to the development of K-Models, a new clustering approach tailored for functional data with an underlying ordinal structure. These curves were previously analyzed in \cite{Patane2026} in a supervised setting with nominal labels; the present work extends that analysis to an unsupervised framework, explicitly accounting for the ordinal structure of the data. The proposed method groups functions based on a common latent effect, distinguishing clusters by the intensity with which this effect manifests. Additionally, it provides estimates of the underlying function and cluster-specific coefficients, offering a more structured and interpretable representation of the data. \\
This study is so organised as follows: in Section \ref{subsec:competitors} we review well-established techniques used for comparison. Section \ref{sec:mtd} introduces the problem and presents our novel clustering method. Section \ref{sec:validation} describes the validation metrics employed to assess clustering performance. Section \ref{sec:simstudy} applies the methods to simulated datasets using a Monte Carlo approach. Finally, in Section \ref{sec:casestudy} we evaluate the methods on a real-world datasets focusing specifically on biosensor signals, the primary motivation for the development of K-Models.

\section{State-of-the-art methods}
\label{subsec:competitors}
The key idea underlying our approach is that the observed functional data are generated by a latent functional phenomenon whose intensity varies progressively across groups, inducing a natural ordering among clusters. A wide range of clustering methods have been developed for functional data, primarily aimed at partitioning curves into homogeneous groups. Among these, we consider three widely used approaches that represent standard benchmarks in the functional clustering literature: K-Means, which partitions data by minimizing within-cluster variance; Partitioning Around Medoids (PAM), a medoid-based alternative more robust to outliers; and Functional Principal Component Analysis (fPCA) combined with K-Means, which reduces data dimensionality before clustering. These methods are well-established, interpretable, and flexible enough to be applied across a broad range of functional data settings. Unlike our method, however, they do not incorporate any ordinal structure among clusters, nor do they provide a direct characterisation of each group in terms of the underlying functional shape.

Formally, we consider a set of functional observations $\{x_i(t)\}_{i=1}^n$, where each function takes values in $\mathbb{R}$ and is defined on a common domain $\mathcal{T}\subset\mathbb{R}$. Given a predefined number of clusters $K$, the objective is to find an optimal partition $\mathcal{C}^*=\{\mathcal{C}_1^*,\dots, 
\mathcal{C}_K^*\}$ that divides the curves into $K$ ordered groups. We assume that the clusters admit a natural ordering $\mathcal{C}_1 < \dots < \mathcal{C}_K$, induced by the progressive amplification of a shared functional effect across levels: each successive cluster exhibits a stronger expression of this component, so that the ordering reflects an intrinsic gradation in the underlying phenomenon rather than an arbitrary labelling.

\subsection{Functional K-Means}
\label{subsec:kmeans}
K-Means is one of the most widely used clustering algorithm. As highlighted by \cite{Bock2007}, its earliest formulation dates back to 1950, although the formal definition was established later.
Over the years, various adaptations of K-Means have been proposed, but they all share a common underlying principle: maximizing the within-cluster similarity between data points and their corresponding cluster centroids, where a centroid is defined as a representative element of the group. This is achieved through an iterative procedure that assigns each data point to the cluster with the closest centroid according to a predefined distance metric. The process ultimately seeks to minimise the within-cluster variance, thereby producing an optimal partition of the dataset. \\
A rigorous mathematical formulation is provided by \cite{Bock2007}, which can be adapted to functional setting by solving the following minimisation problem:
\begin{equation}
\label{eq:kmeans_objective}
     (\mathcal{C}^*, \mathcal{Z}^*) = \arg\min_{\mathcal{C}, \mathcal{Z}} \sum_{k=1}^K \sum_{x_i \in C_k} d^2(x_i, z_k),
\end{equation}

where $\mathcal{C}$ is the clustering partition, $K$ the number of clusters, $\mathcal{Z}=\{z_1, \dots, z_K\}$ represents the set of centroids, and $d(\cdot,\cdot)$ is a measure of dissimilarity (or equivalently, similarity) between two functions. Moreover, to mitigate the sensitivity to initialisation, a multiple-start strategy is commonly employed. From Equation~\eqref{eq:kmeans_objective} it is evident that K-Means is a highly flexible approach since it imposes no strict assumptions on the data structure and allows for the use of different dissimilarity measures $d$, making it adaptable to various contexts.\\
When dealing with functional data, one of the most commonly used metrics, and the one adopted in this study, is the $L^2$-norm -- i.e. for $x\in L^2(I)$, $||x||_{L^2}=\int_I x(t) dt$ -- :
\begin{equation}
    \label{eq:kmeans_metric}
        d(x_i, z_k):=\| x_i - z_k \|_{L^2}. 
    \end{equation}

This measure ensures that clustering is based on the overall shape of the functions rather than specific pointwise differences, making it a suitable choice for functional data.\\
Despite its conceptual simplicity, functional K-Means inherits several well-known limitations from its multivariate counterpart, including sensitivity to initialisation and an implicit assumption of "spherical" clusters.

\subsection{Functional PAM}
\label{subsec:pam}
Partitioning Around Medoids (PAM, \cite{Kaufman1990}), also known as K-Medoids, is another widely used clustering algorithm. While its core idea is similar to that of K-Means, a key distinction lies in the definition of the representative element of a cluster.
As discussed in Section~\ref{subsec:kmeans}, K-Means defines centroids as the mean of all elements within a cluster. While this is an effective representative in Euclidean spaces, it may not be the most suitable choice for other dissimilarity measures.
\\
In contrast, PAM represents each cluster with a medoid, which is the data point that minimises the average (or equivalently, the total sum of) dissimilarities to all other points in the same cluster. Formally, given a cluster $C_k$, we define its medoid as:

\begin{equation}
\label{eq:medoid def}
   m_k:=\arg\min_{x_i \in C_k}\sum_{x_j \in C_k}d(x_i, x_j).
\end{equation}

This formulation highlights a key difference from K-Means:  while a centroid $z_k$ can be any point in the data space, a medoid $m_k$ must be an actual data point from the dataset. This property makes K-Medoids more robust to outliers and allows its application to any arbitrary dissimilarity measures.\\
Similar to K-Means, the K-Medoids method relies on minimizing an absolute error criterion, commonly referred to as total deviation ($TD$), which is defined as:

\begin{equation}
\label{eq:tot deviation}
    TD := \sum_{k=1}^K \sum_{x_i \in C_k} d(x_i, m_k),
\end{equation}

where $d(\cdot,\cdot)$ is a dissimilarity measure, while $\{m_1, \dots,m_K\}$ represents the set of medoids for the partition $\{C_1,\dots,C_K\}$.\\
In this study, the PAM algorithm is implemented by using the $L^2$-norm as dissimilarity measure for functional data. By adopting the $L^2$-norm, the method ensures a clustering structure that reflects meaningful variations in the underlying functional patterns while preserving the robustness of the PAM framework.

\subsection{Functional PCA combined with K-Means}
\label{subsec:fpca_kmean}
FPCA combined with K-Means -- drawing inspiration from~\cite{Peng_2008} and~\cite{Wu18102022} -- first reduces the dimensionality of the functional dataset using Functional Principal Component Analysis (fPCA), and only then performs the clustering. A key advantage of this method is that it allows clustering to be performed directly in the directions of greatest variability in the dataset, potentially leading to more meaningful partitions.
\BlankLine
Let $X$ be a stochastic process in $L^2(\mathcal{T)}$. By exploiting the Karhunen-Loève expansion, $X$ can be expressed as:
\begin{equation}
\label{eq:KL expansion}
    X(t) = \mu(t) + \sum_{l=1}^{\infty} \alpha_l \phi_l(t) \quad \forall t \in \mathcal{T},
\end{equation}
where $\mu(t):=\mathbb{E}[X(t)]$ is the mean function, $\{\phi_l\}_{l=1}^{\infty}$ are the principal functions and $\{\alpha_l\}_{l=1}^{\infty}$ the corresponding principal component scores, defined as:
\begin{equation}
\label{eq:fPCA scores}
    \alpha_l:=\int_{\mathcal{T}} (X(t)-\mu(t)) \phi_l \, dt \quad l=1,2,...
\end{equation}
Retaining only the scores associated with the first $p$, such that the cumulative proportion of variance retained is, for instance, at least $95\%$ principal components, we obtain a new multivariate dataset of dimensions $n \times p$, where each curve is now represented by the corresponding vector of scores, $\{\boldsymbol{\alpha}_i \}_{i=1}^{n}$, with $\boldsymbol{\alpha}_i \in \mathbb{R}^p$.

At this point, clustering is performed by applying the classical multivariate K-Means. The only modification lies in the choice of the dissimilarity measure: instead of using the $L^2$-norm between functional observations, clustering is now based on the Euclidean distance between the extracted score vectors. This reformulation simplifies the problem by shifting from an infinite-dimensional functional space to a finite-dimensional multivariate setting, while still capturing the dominant modes of variability in the data.

\section{K-Models}
\label{sec:mtd}

The aim of this work is to propose a new method for the unsupervised clustering of functional data. In general, clustering seeks to partition observations into distinct subgroups such that instances within the same group are highly similar, while those belonging to different groups are as dissimilar as possible. 

In the context of functional data, clustering is particularly useful for identifying homogeneous and heterogeneous patterns among curves and for revealing potential structures in the underlying stochastic process. The notion of clustering's optimality depends on the clustering criterion adopted, which varies across methods according to the type of similarity (or dissimilarity) being emphasised.

Most existing clustering methods for functional data focus on identifying groups that differ in shape or location, treating clusters as nominal categories without any inherent structure between them. In contrast, the method proposed in this work introduces an additional structural assumption: clusters are not only distinct but also \textit{ordered}. In other words, we aim to perform \textit{ordinal clustering}, where the clusters represent increasing levels of a latent functional effect.


The objective of the method is therefore twofold: (i) to estimate the latent functional direction capturing the shared phenomenon, and (ii) to partition the curves according to the degree to which they amplify this phenomenon. Clusters thus correspond to different \textit{levels} of the same underlying functional effect rather than to unrelated groups of curves. This interpretation naturally yields an ordered structure among clusters.

Under this framework, we consider the following cluster-specific functional model:
\begin{equation}
    \label{eq:k-mod}
    x_i(t) = \mu_0(t) + \mu_{k}(t) + \varsigma_k\,\beta(t) + \varepsilon_i(t), 
    \qquad t \in \mathcal{T},
\end{equation}
subject to $\|\beta\|_{L^2}=1$ and $\langle \mu_k, \beta \rangle = 0$, where $x_i$ belongs to cluster $\mathcal{C}_k$, with $k\in \{1,\dots,K\}$. Here, $\mu_0(t)$ denotes the global mean function, $\mu_k(t)$ represents the deviation of cluster $\mathcal{C}_k$ from the global mean, and $\varsigma_k$ is a scalar coefficient measuring the amplitude of the shared functional direction $\beta(t)$ within cluster $\mathcal{C}_k$. The function $\beta(t)$ captures a common functional component that is expressed with different intensities across clusters, while $\varepsilon_i(t)$ is a zero-mean random noise process.

The ordinal structure arises naturally from this decomposition: we assume $\mathcal{C}_i < \mathcal{C}_j$ whenever $\varsigma_i < \varsigma_j$, so that the ordering among clusters is fully encoded in the scalar coefficients $\varsigma_k$. The constraint $\|\beta\|_{L^2}=1$ fixes the scale of the ordinal component, while 
$\langle \mu_k, \beta \rangle = 0$ ensures a unique decomposition into an ordinal part $\varsigma_k\beta$ and a cluster-specific deviation $\mu_k$.

The ordinal nature of the latent classes is determined by the values of the coefficients $\varsigma_k$, which induce a hierarchical ordering of the clusters according to increasing levels of amplification of $\beta(t)$. There exist two equivalent orderings, corresponding to the two possible sign choices of $\beta$ (and $\varsigma_k$); these lead to reversed but otherwise identical interpretations of the clustering structure. In practice, one can select the orientation that best supports scientific communication and interpretability.

Therefore, unlike conventional clustering methods that simply partition the data into unrelated groups, our approach yields clusters that are intrinsically ordered, providing a natural and interpretable ranking of the curves according to the strength of the latent functional effect. The clustering procedure is formulated as an optimisation problem aimed at minimizing the within-cluster variability along the latent ordinal direction. Specifically, we define the objective function as
\begin{equation}
\label{eq:kmod_loss}
\mathcal{L}((\mathbf{\varsigma}, \boldsymbol{\beta}), \mathcal{C}) 
= \sum_{k=1}^{K} \sum_{i \in C_k} 
\left(\langle x_i-\mu_0, \beta\rangle - \varsigma_k\right)^2
+\lambda\,\mathcal{J}(\beta).
\end{equation}
where the component $\mu_k$ does not appear due to the constraint $\langle\mu_k,\beta\rangle=0$, and where the term $\mathcal{J}(\cdot)$ is a roughness penalty, ensuring that the estimated latent functions remain smooth and interpretable. In this study, we impose the penalty on the integral of the second derivative; namely, given a function $\gamma\in L^2$:
\begin{equation}
\label{eq:penalty J}
    \mathcal{J}(\gamma) = \int_{\mathcal{T}} \left( \gamma''(s) \right)^2 ds.
\end{equation}
This loss function measures the within-cluster dispersion of the projections of the curves onto the direction $\beta$. By minimizing this quantity, the method identifies clusters whose members exhibit similar levels of amplification of the latent functional effect. Consequently, the procedure performs \textit{ordinal clustering}, organizing the functional observations along a latent ordered dimension, rather than producing a generic partition of the data as in classical approaches such as functional $K$-means.

The regularisation parameter $\lambda$ controls the trade-off between model fit and smoothness. A higher value of $\lambda$ enforces stronger regularisation, favoring smoother estimates of the functional parameters, at the cost of a potentially higher approximation error within clusters. 

By minimizing this objective function, the method achieves a partitioning of the curves where each group exhibits a different level of the underlying phenomenon, while ensuring that $\beta$ remains a smooth and well-structured function. This approach leads to an interpretable clustering solution that reflects a natural hierarchical ordering among the identified subgroups.
\subsection{Implementation}
\label{subsec:K-mod_implementation}
The proposed clustering procedure iteratively alternates between the following two key steps:
\begin{enumerate}
    \item Estimation of the functions $\mu_k(t)$ and quantity $\varsigma_{k}$ for each cluster independently, while the common functions $\mu_0(t),\beta(t)$ are calculated using the entire data set.
    \item Updating the cluster assignments by grouping together curves that exhibit similar patterns, based on the estimated coefficients.
\end{enumerate} 
A former approach, ~\cite{FRECL}, proposes a clustering algorithm for functional data based on cluster-specific functional regression models. Our method differs in a fundamental aspect: in ~\cite{FRECL}, the observed curves $x_i$ are modeled through a set of predefined functional covariates \[\boldsymbol{X}_{i}(t)=\begin{bmatrix} X_{i1}(t), \dots ,  X_{iL}(t)\end{bmatrix}^\top\]
$t\in\mathcal{T}$, which influence the estimation process. In contrast, our approach does not rely on any external covariates. Instead, we estimate both the latent function $\beta(t)$ and the cluster-specific parameters $(\mu_k(t),\varsigma_k)$ directly from the data, without any additional input variables.  This ensures that the clustering is driven purely by the intrinsic shape of the observed curves.

Algorithm \ref{alg:kmodels} implements the proposed K-Models procedure for ordinal clustering of functional data. Each functional observation $x_i(t)$ and the latent direction $\beta(t)$ are represented in a common basis expansion
\[
x_i(t) = \sum_{j=1}^{n_J} a_{ij}\phi_j(t), 
\qquad
\beta(t) = \sum_{j=1}^{n_J} b_j \phi_j(t),
\]
where $a_i=(a_{i1},\dots,a_{in_{J}})^\top$ and $b=(b_1,\dots,b_{n_J})^\top$ denote the corresponding coefficient vectors and $\{\phi_j(t)\}_{j=1}^{n_J}$ is a chosen functional basis. Coherently with this finite structure, we define two useful matrices:
\begin{itemize}
    \item $D_2=\left\{\int_{I}\frac{\partial^2 \phi_j(t)}{\partial t^2}\frac{\partial^2 \phi_k(t)}{\partial t^2}\,dt\right\}_{j,k}$
    \item $D_0=\left\{\int_{I}\phi_j(t)\phi_k(t)\,dt\right\}_{j,k}$
\end{itemize}
This representation allows all computations to be carried out in the finite-dimensional coefficient space, while preserving the functional nature of the data.

We initialise the clusters randomly, to enable exploration of possible clustering configurations, and $\beta(t)$ through fPCA, to suggest a plausible quantitative -- ideally ordinal -- direction. The algorithm alternates between three main steps in each iteration. First, given the current latent direction $\beta(t)$, cluster-specific parameters are estimated: the cluster means in the coefficient space $\bar{a}_k$, the cluster-specific intercepts $\beta_{0k}$, and the ordinal amplification coefficients $\varsigma_k$. The amplification coefficient $\varsigma_k$ represents the strength with which the common latent effect $\beta(t)$ manifests in cluster $k$. 

Second, the latent direction $\beta(t)$ is updated using the current cluster assignments and amplification coefficients. The update Equation \eqref{eq:update_beta} is proved in Appendix \ref{appendix:update_beta}. This step ensures that $\beta(t)$ captures the direction of variation associated with the ordinal latent effect.

Third, curves are reassigned to clusters based on the proximity of their projection onto $\beta(t)$ to the current cluster levels $\varsigma_k$. Specifically, the projection score of each curve is compared to each cluster’s amplification coefficient, and the curve is assigned to the cluster minimizing the squared distance. After reassignment, clusters are reordered to enforce the ordinal structure $\varsigma_1 \le \dots \le \varsigma_K$.

To guarantee identifiability, $\beta$ is $L^2$-normalised by imposing $\|\boldsymbol b\|_{D_0}=1$.

To prevent empty or degenerate clusters, a repopulation safeguard is applied: if any cluster contains fewer than a minimum number of curves, a small number of curves are randomly reassigned to that cluster. This mechanism ensures stability of the alternating optimisation and avoids convergence to trivial solutions.

Finally, the procedure can be combined with a multi-start strategy to improve robustness. Multiple runs with random initialisations are performed, and the clustering solution with the highest quality score (e.g., a ratio of between-cluster to within-cluster variation) is retained as the final output. The algorithm iterates until the cluster assignments converge, producing the latent function $\beta(t)$, the ordered cluster amplification coefficients $\{\varsigma_k\}$, and the final cluster partition $\mathcal{C}$.

\newpage
\begin{algorithm}[H]
\footnotesize
\caption{K-Models: Ordinal Functional Clustering}
\label{alg:kmodels}
\textbf{Input:} Functional data $\{x_i(t)\}_{i=1}^n$, number of clusters $K$, maximum iterations $I_{\max}$  

\textbf{Basis representation:} Let $\{\phi_j(t)\}_{j=1}^{n_J}$ be a basis of the functional space. \[ x_i(t)=\sum_{j=1}^{n_J} a_{ij}\phi_j(t), \qquad \beta(t)=\sum_{j=1}^{n_J} b_j\phi_j(t), \qquad \mu_k(t)=\sum_{j=1}^{n_J} c_{kj}\phi_j(t) \] Denote by \[ \mathbf{a}_i=(a_{i1},\dots,a_{in_J})^\top, \qquad \mathbf{b}=(b_1,\dots,b_{n_J})^\top, \qquad \mathbf{c}_k=(c_{k1},\dots,c_{kn_J})^\top \] the corresponding coefficient vectors.

\textbf{Initialisation:}  

Center the data: $x_i(t)\leftarrow x_i(t)-(\frac{1}{n}\sum_{i=1}^n x_i(t))$. Save $\mu_0=(\frac{1}{n}\sum_{i=1}^n x_i(t))$.

Initialise clusters $\{C_k\}$ randomly and estimate $\beta(t)$ using the first fPCA component.

\For{$it = 1,\dots,I_{\max}$}{

\textbf{Cluster update:}

\For{$k=1,\dots,K$}{
Compute cluster mean $\bar{\mathbf{a}}_k=\frac{1}{n_k}\sum_{i\in \mathcal{C}_k} \boldsymbol{a_i}$, where $n_k=|\mathcal{C}_k|$ for any $k\in\{1,...,K\}$.

Compute cluster amplitude of the effect $\varsigma_k \leftarrow \bar{\mathbf{a}}_k^\top D_0 \mathbf{b}$\; 
$\mathbf{c}_k \leftarrow \bar{\mathbf{a}}_k - \varsigma_k \mathbf{b}$\;
}

\textbf{Latent direction update:}

Find $\mathbf{b}$ such that:

\begin{equation}
\label{eq:update_beta}
\bigg(\sum_{k=1}^{K} n_k \varsigma_k^2 D_0 + \lambda D_2\bigg)\mathbf{b}
\leftarrow 
\sum_{k=1}^{K} n_k \varsigma_k (\bar{\mathbf{a}}_k-\mathbf{c}_k)
\end{equation}

\textbf{Ordinal enforcement:}

Order clusters such that $\varsigma_1 \le \dots \le \varsigma_K$.

\textbf{Reassignment:}

\For{$i=1,\dots,n$}{
$\rho_i \leftarrow \dfrac{\mathbf{a}_i^\top D_0 \mathbf{b}}{\mathbf{b}^\top D_0 \mathbf{b}}$; \;
$C_i \leftarrow \arg\min_k (\rho_i-\varsigma_k)^2$
}

\textbf{Normalisation for identifiability:}

\[
\mathbf{b} \leftarrow \mathbf{b}\cdot (\mathbf{b}^\top D_0 \mathbf{b})^{-1}
\]

\textbf{Repopulation safeguard:}

If any cluster contains fewer than a minimum number of curves, randomly reassign a small set of observations to that cluster.

\textbf{Convergence check:}

Stop if cluster assignments do not change.
}

\textbf{Output:} $\beta(t)$, $\{\varsigma_k\}$, cluster partition $\mathcal{C}$

\end{algorithm}

\section{Validation in clustering}
\label{sec:validation}
To assess the performance of the proposed clustering methods, we employ two widely used evaluation metrics: the Between-Cluster to Within-Cluster Variability Ratio and the Silhouette Score. These two measures provide complementary perspectives on clustering quality, offering both a global and a local evaluation of the resulting partitions. \\
The first metric quantifies how well-separated the clusters are relative to their internal cohesion. It is defined as the ratio between the Between-Cluster Sum of Squares (BSS) and the Within-Cluster Sum of Squares (WSS). Given a partition $\mathcal{C}= \{C_1, \dots, C_K\}$ of the functional observations, the ratio is computed as follows:
\begin{equation}
\label{eq:ratio}
\begin{aligned}
\text{Ratio} &= \frac{\text{BSS}}{\text{WSS}}, \quad \text{with} \\
\text{BSS} &:= \sum_{k=1}^{K} n_k \| \mu_k - \mu_0 \|^2_{L^2}, \\
\text{WSS} &:= \sum_{k=1}^{K} \sum_{x_i \in C_k} \| x_i - \mu_k \|^2_{L^2}.
\end{aligned}
\end{equation}
Where $n_k = |C_k|$ denotes the number of curves in cluster $k$, $\mu_k(t) := \frac{1}{n_k} \sum_{x_i \in C_k} x_i(t)$ is the pointwise mean of the functions within cluster $C_k$, and $\mu_0(t) := \frac{1}{n} \sum_{i=1}^{n} x_i(t)$ is the global pointwise mean. A higher BSS/WSS ratio indicates well-separated clusters with high internal homogeneity, which are desirable properties in clustering. \\
The second metric used is the Silhouette Score. As described in~\cite{SSscore}, it evaluates clustering quality by balancing intra-cluster cohesion and inter-cluster separation. For each observation $x_i$, the Silhouette Score is defined as:
\begin{equation}
\label{eq:ss score}
s_i := \frac{d_i - w_i}{\max(d_i, w_i)}, \quad i=1,\dots,n.
\end{equation}
where $w_i$ is the average $L^2$-distance between $x_i$ and all other curves within the same cluster (a measure of intra-cluster similarity), while $d_i$ is the smallest average $L^2$-distance between $x_i$ and all curves in a different cluster (a measure of inter-cluster separation). Then the overall Silhouette Score is computed as the mean of $s_i$ across all observations:
\begin{equation}
\label{eq: overall ss score}
S=\frac{1}{n}\sum_{i=1}^{n}s_i.
\end{equation}

This score ranges between -1 and 1, with higher values corresponding to better partitions of the curves.
\BlankLine
These two metrics evaluate clustering quality from different but complementary perspectives:
\begin{itemize}
    \item \textbf{BSS/WSS Ratio}: provides a global measure of clustering performance by comparing the between-cluster sum of squares (BSS) to the within-cluster sum of squares (WSS). A higher ratio indicates better separation between clusters relative to their internal variability. This metric is naturally optimised by most clustering methods, but in multiple-start procedures it is also used to select the best solution among different initialisations.
    
    \item \textbf{Silhouette Score}: captures local clustering behavior by assessing how well each individual curve is assigned to its respective cluster. Specifically, it compares the average distance of a curve to other curves in the same cluster with the average distance to curves in the nearest neighboring cluster. This metric is particularly useful to identify potentially misclassified curves that lie near cluster boundaries and to compare the quality of different clustering methodologies.
\end{itemize}
\BlankLine
By combining these two evaluation approaches, the BSS/WSS ratio ensures that clusters are compact and well-separated on a global scale, while the Silhouette Score validates the assignment consistency at an individual level. This dual perspective leads to a more robust assessment of the clustering solutions in functional data analysis. 

Now, we present a simulation study designed to validate the K-models algorithm. The aim of the simulation study is to evaluate the ability of the method to correctly recover the underlying ordinal clustering structure under controlled settings. In particular, synthetic functional datasets are generated with known ordinal patterns and varying levels of noise and cluster separation, allowing us to systematically investigate the robustness and accuracy of the proposed method.

\section{Simulation Study}
\label{sec:simstudy}
To evaluate the empirical performance of the considered clustering approaches, we conduct a simulation study based on synthetic functional data generated from a realistic structure derived from the dataset of our case study. The aim of the study is to assess how accurately the different methods recover the underlying clustering structure under varying proportions of ordinal and global variability.

The simulated data are generated according to a functional model constructed from the empirical functional principal components of the original dataset. In particular, the mean function, the first principal component (PC1), and the second principal component (PC2) are taken directly from the case study. The second component is used to generate a global smooth variability shared across all observations, whereas the first component is used to introduce the cluster structure.

More precisely, let $K$ denote the number of clusters and $n$ the number of functional observations. Each curve is generated as a perturbation of the empirical mean function plus a linear combination of the first two functional principal components. The score associated with the second component is drawn from a Gaussian distribution representing global functional variability common to all curves. In contrast, the score associated with the first component is generated from cluster-specific Gaussian distributions whose means determine the cluster centers. Consequently, clusters differ primarily along the first principal direction.

\begin{figure}[H]
    \centering
    \includegraphics[width=1\linewidth]{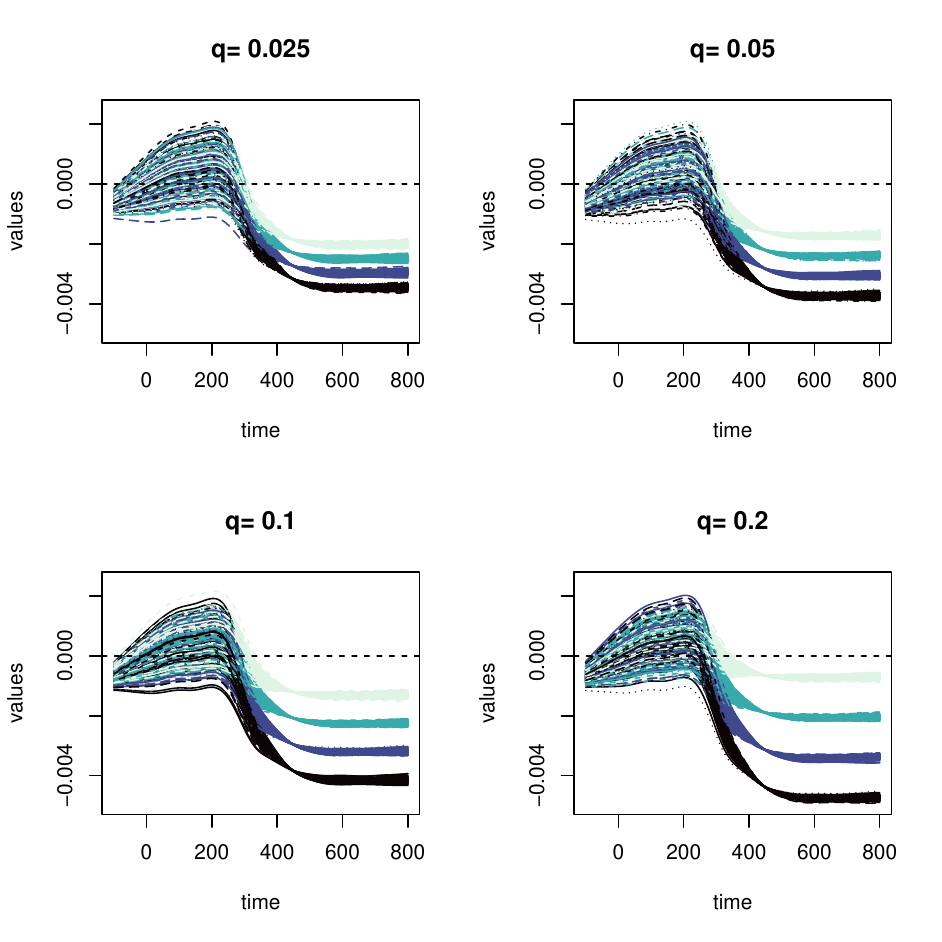}
    \caption{Examples of simulated scenarios depending on $q$.}
    \label{fig:examplesim_kmodels}
\end{figure}

To control the relative importance of the cluster structure with respect to the global variability, we introduce a parameter $q$ that scales the variability associated with the first principal component. This parameter regulates the ratio between the ordinal (cluster-related) variability and the global variability while maintaining a clear separation between clusters. In the simulation study we consider the values
\[
q\in \{0.025, 0.05, 0.1, 0.2\}.
\]
As an illustrative example Figure \ref{fig:examplesim_kmodels} displays one example of simulated dataset per considered value of $q$.

For each configuration we generate multiple Monte Carlo samples and apply the clustering methods under comparison. Since the true cluster labels are known by construction, the performance of each method is evaluated by measuring the agreement between the estimated partition and the true clustering structure. The clusters are constructed to guarantee a clear separation between groups, so that the true partition is well defined and not subject to ambiguity. 

\begin{figure}[H]
    \centering
    \includegraphics[width=0.8\linewidth]{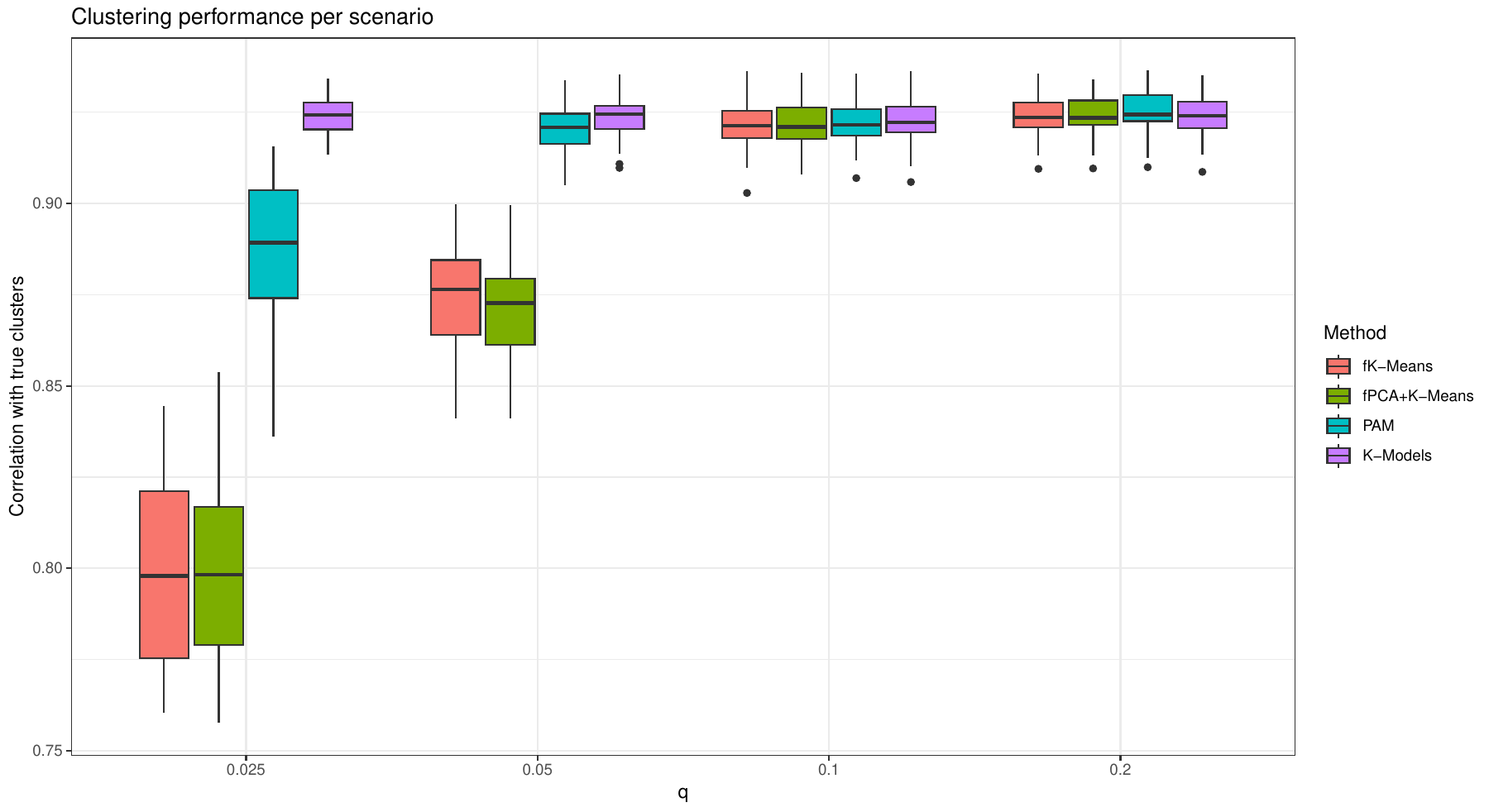}
    \caption{Boxplots of the correlations obtained across simulation runs for each value of $q$ and for each clustering method.}
    \label{fig:results_sim_kmodels}
\end{figure}

To quantify the agreement between the true and the estimated partitions, we compute the correlation between the true cluster labels and the estimated labels. In the case of K-models, the clusters are naturally ordered, enabling a direct computation of the correlation. For our competitors, we allow for label switching, considering all possible permutations of the estimated labels and retain the maximum correlation over these permutations. This ensures that the measure is invariant to the arbitrary labeling of clusters, and ensures a fair comparison with K-models. In Equation \ref{correlation_clust}, we define the correlation estimated $\hat{\mathcal{C}}$ and true $\mathcal{C}$ clustering configuration, for generic clustering methods.

\begin{equation}
\operatorname*{Corr}(\hat{\mathcal{C}},\mathcal{C})=\max_{\pi \in \operatorname{Sym}*(K)}
\;\operatorname*{Corr}(k_i, \pi(\hat k_i))
\label{correlation_clust}
\end{equation}
where $k_i=k\cdot\mathbbm{1}_{i\in C_k} $, and $\operatorname{Sym}(K)$ is the group of permutations on $\{1,...,K\}$. While alternative measures such as the Adjusted Rand Index (ARI) could also be used, we prefer this criterion because, in addition to being permutation invariant, it provides a more stringent notion of agreement. Indeed, by relying on the numerical values of the labels after optimal alignment, it captures how well the overall labeling structure matches the true one, rather than only pairwise group membership. This is particularly relevant in our simulation setting, where clusters are generated along a continuous latent direction. 
In Figure \ref{fig:results_sim_kmodels}, the results are summarised using boxplots of the correlations obtained across simulation runs for each value of $q$ and for each clustering method.  This representation allows us to analyze how the performance of the competing approaches varies as the relative contribution of ordinal variability changes. In particular, K-Models approach consistently achieves high correlations even when $q$ is small, indicating its ability to detect subtle cluster structures. By contrast, competing methods show more variability and generally lower correlations in these challenging scenarios. As $q$ increases, the differences between methods tend to diminish, but the proposed approach maintains a robust performance across all considered values of $q$, illustrating its stability and reliability in recovering the true clustering structure.

\section{A case study: ordinal clustering of optical biosensor signals}
\label{sec:casestudy}

In this section, we apply the proposed method to our case study, based on biosensor signals. 
The signals are treated as functional data obtained through the preprocessing procedure described in \cite{Patane2026}, where the same dataset is analyzed in a supervised setting.

\begin{figure}[H]
    \centering
    \hspace{15mm}
    \includegraphics[width=0.9\textwidth]{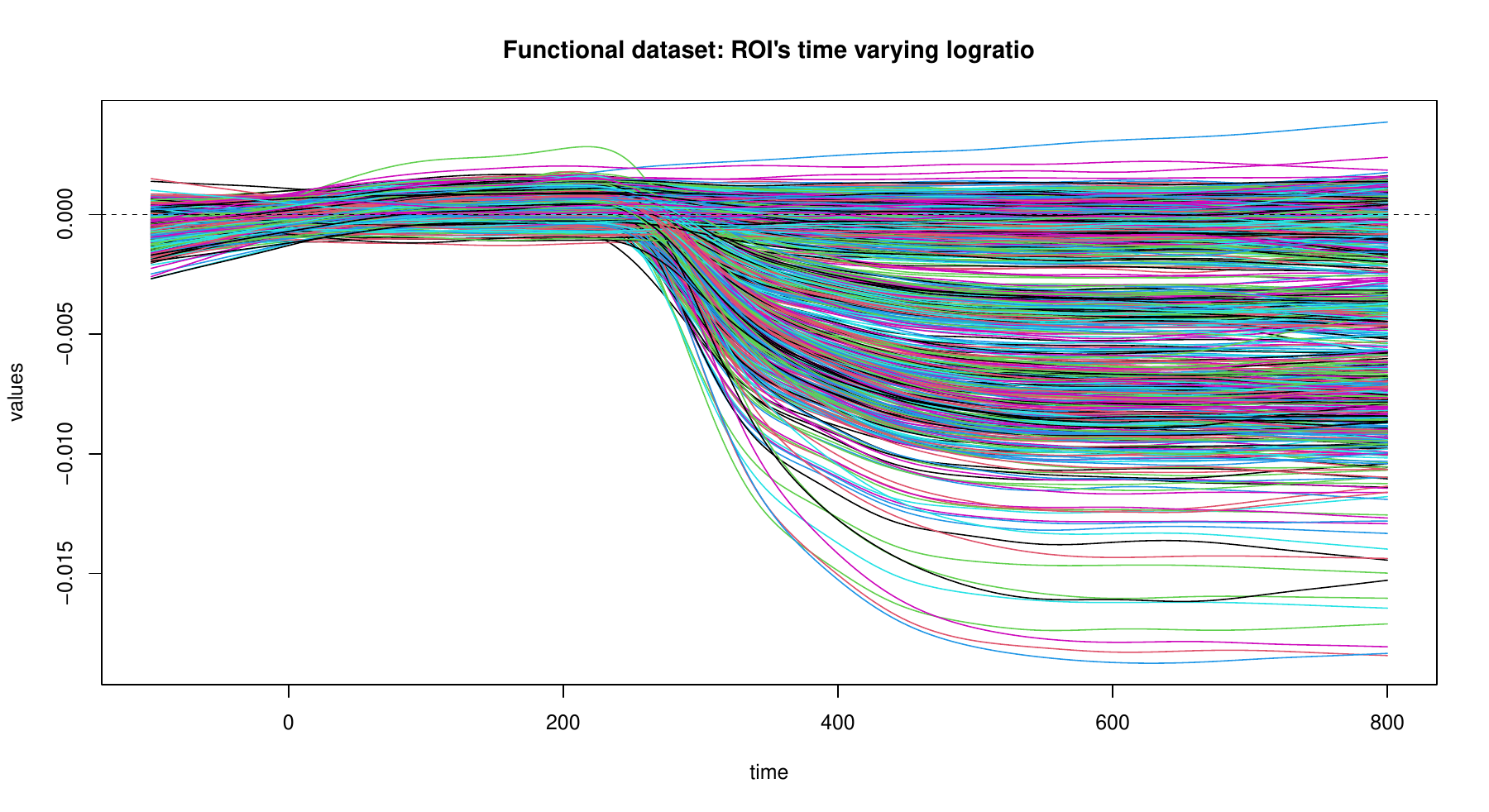}
    \caption{Unlabeled ROI signal curves.}
    \label{fig:roi_curves}
\end{figure}

The functional dataset, displayed in Figure~\ref{fig:roi_curves}, consists of biosensor signal profiles from $1026$ Regions of Interest (ROIs), i.e., there are $1026$ spots with eight different antigen concentrations immobilized.
Each signal represents the temporal evolution of light intensity within the corresponding ROI. 
As already discussed in \cite{Patane2026}, using an optical reflectometric sensor, light intensity is used as an indicator of interactions between antigenes and antibodies \cite{gauglitz10}.

\begin{figure}[H]
    \centering
    \hspace{15mm}
    \includegraphics[width=0.9\textwidth]{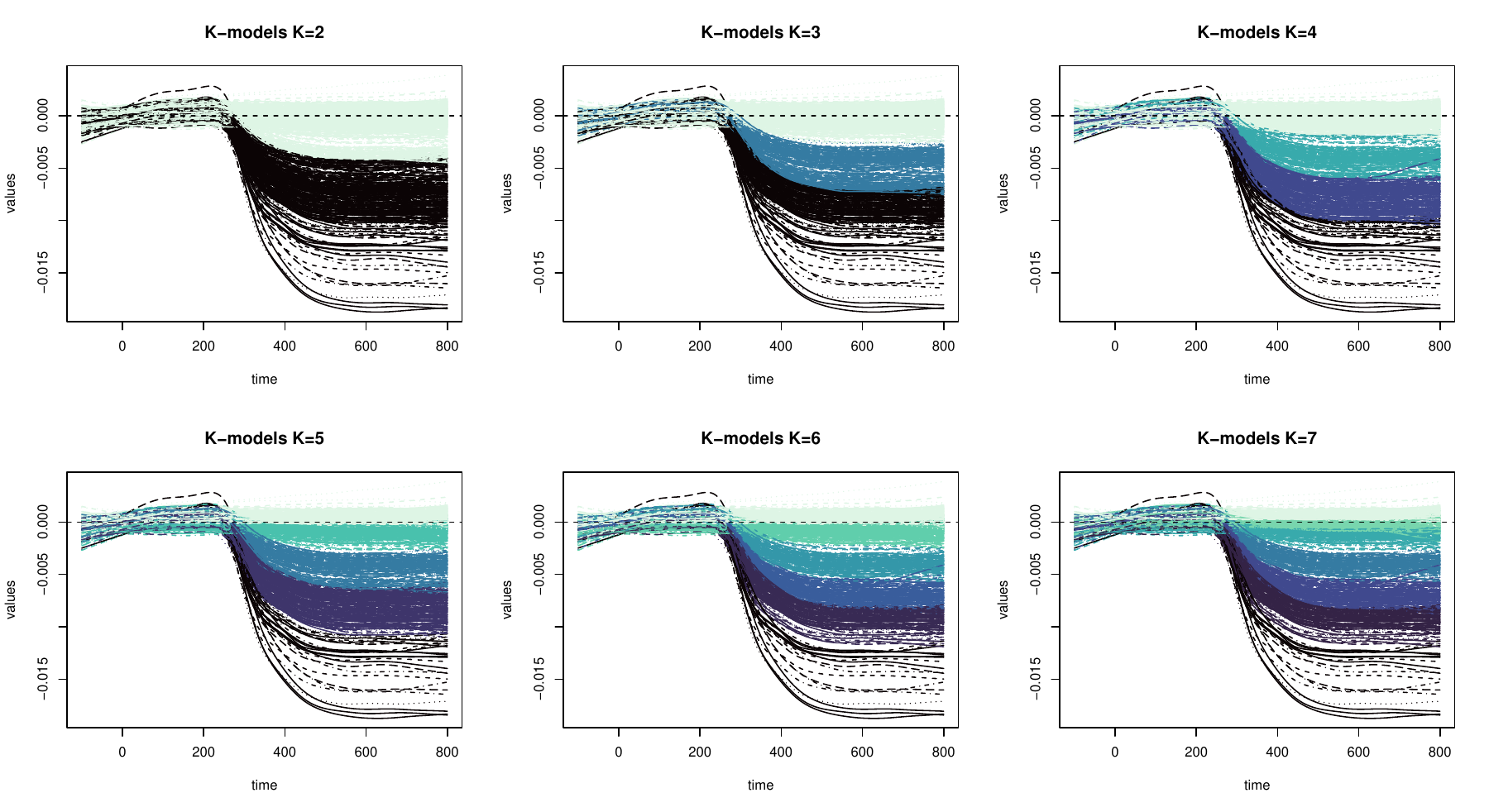}
    \caption{The levels assigned to the biosensor signals by the K-models $K=2,...,7$.}
    \label{fig:comparison_roi}
\end{figure}

\begin{figure}[H]
    \centering
    \includegraphics[width=0.8\linewidth]{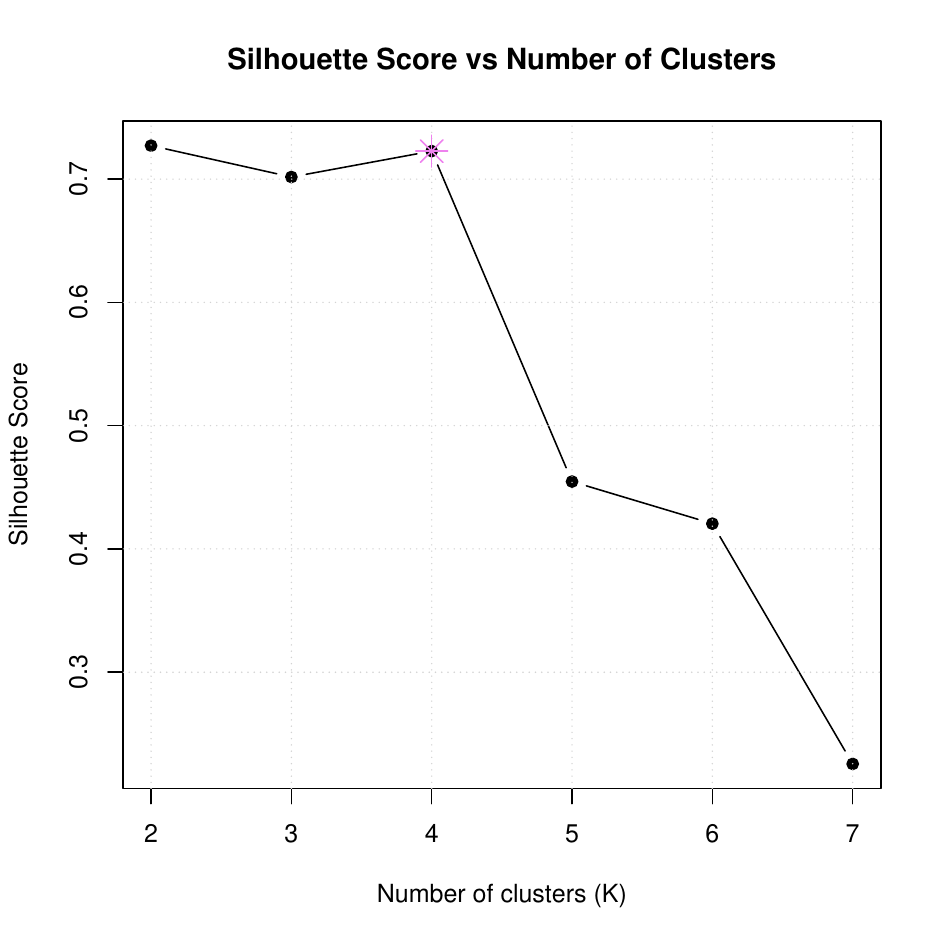}
    \caption{Silhouette score of the projections of $x_i$ on $\beta$, by the K-models $K=2,...,7$. We choose $K=4$ according the elbow rule.}
    \label{fig:sihlouette_casestudy}
\end{figure}

The K-models algorithm is applied to the biosensor signals by considering a number of clusters $K$ ranging from $2$ to $7$. Figure~\ref{fig:comparison_roi} illustrates the resulting partitions of the dataset for different values of $K$. In this real-data application, both the underlying ordinal structure and the appropriate number of clusters are unknown. Therefore, model selection is performed using the Silhouette Score. Indeed, since the algorithm already maximises the between-cluster to within-cluster sum of squares ratio (BSS/WSS), the Silhouette Score provides a complementary criterion for assessing clustering quality. According to the Silhouette Score plot reported in Figure~\ref{fig:sihlouette_casestudy}, the elbow rule suggests selecting $K=4$ as the optimal number of clusters, fully signal-based. This differs from the 8 nominal concentration levels known from the experimental design -- used to supervise the ordinal classification in \cite{Patane2026} -- indicating that the light intensity signal alone is not sufficient to discriminate all antigen concentrations, as some levels appear indistinguishable from a purely signal-based perspective.

\begin{figure}[H]
    \centering
    \hspace{15mm}
    \includegraphics[width=1\textwidth]{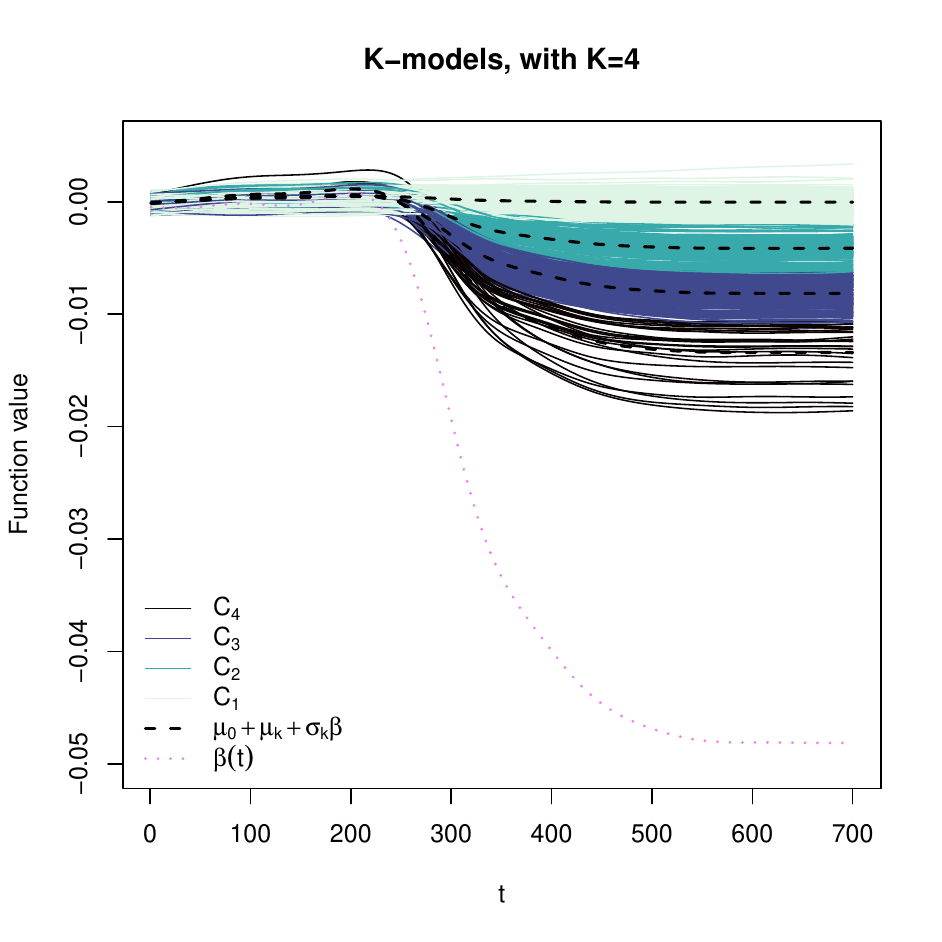}
    \caption{The ordinal pattern $\beta$ which increases with $k$ according to $\varsigma_k$.}
    \label{fig:beta1_roi}
\end{figure}

Once $K$ is fixed, the model outputs can be more fully leveraged for interpretation. In particular, the $K$-models provide an estimate of the ordinal behavior $\beta$ (Figure~\ref{fig:beta1_roi}). Moreover, the clusters identified by the method are naturally ordered according to the values of $\varsigma_k$, which can be interpreted as different levels of amplification of $\beta$. This ordering---reversible upon changing the sign of $\beta$---is given by $C_1, C_2, C_3, C_4$. Cluster $C_1$ corresponds to the lowest amplitude, likely representing an absent or negligible antibody--antigen interactions. At the opposite extreme, cluster $C_4$ captures a highly pronounced and heterogeneous class of larger signals representing higher antibody-antigen interactions. The remaining clusters represent intermediate levels of response intensity.

As illustrated in Figure~\ref{fig:beta1_roi}, the ordinal behavior $\beta$ clearly characterises the reaction profile: it exhibits a marked decay approximately between time $200$ and $400$, followed by a horizontal asymptote after time $600$. This pattern indicates that the proposed method successfully captures the ordinal latent structure of the data, revealing the underlying level-based organisation of the biosensor signals.

\section{Conclusions}

The great variety of clustering methods for functional data in the existing literature highlights the significant interest that this field has received within the mathematical and engineering communities. Despite this, in several contexts -- such as the biosensor signals presented in the case study -- the existing techniques do not always provide the most suitable tools for addressing the desired analyses. In many cases, clustering procedures are needed not only to partition the curves, but also to offer a comprehension of the underlying mechanisms behind the generation of these functions, and how they contribute to the formation of distinct subgroups.

In this regard, the K-Models method proposed by this work allows for the simultaneous achievement of these two objectives. By defining a clustering framework based on a linear functional model, the K-Models approach enables the estimation of a latent phenomenon underlying the stochastic process that generated the observed curves. Furthermore, it evaluates how this shared effect manifests within each cluster through its corresponding linear coefficient. A key advantage of this formulation is that it defines an ordinal partition of the data, an absent feature in the traditional functional clustering procedures. The ordering of the clusters is indeed directly determined by the magnitude of the common effect, providing a structured interpretation of the underlying relationships within the dataset.

The simulation study presented in this work confirms these theoretical advantages. Across scenarios with subtle cluster-related variability, the K-Models approach consistently recovers the true clustering structure more accurately than competing methods, while maintaining the ordinal interpretation of the clusters. These results illustrate not only the reliability of the method in practical applications, but also its ability to offer a meaningful characterisation of the latent processes driving the observed functional data.

\section*{Acknowledgements}
This research has received funding by the European Commission under the “HORIZON-CL4-2021-DIGITALEMERGING-01 project BioProS - Biointelligent Production Sensor to Measure Viral Activity” (grant agreement no. 101070120), 2022-2026”. GP, AM, LD and FN acknowledge the initiative “Dipartimento di Eccellenza 2023–2027”, MUR, Italy, Dipartimento di Matematica, Politecnico di Milano. LD acknowledges his membership to INdAM GNCS - Gruppo Nazionale per il Calcolo Scientifico (National Group for Scientific Computing), Italy. The authors thank Alessandro Ganelli and Johanna Hutterer for their support in this research.

\section*{Declarations and statements}
\textbf{Conflict of interest}: the authors have no relevant financial or non-financial interests to disclose.

 \appendix

 \label{appendix:update_beta}
\textbf{Proof of Equation \eqref{eq:update_beta}}: in the second step, where $\mu_k$, $\boldsymbol{\varsigma}_k$ and the clusters assignment are fixed, we aim at minimising the conditional loss $\mathcal{L}_c$, where

\[
\mathcal{L}_c(\beta|\mu_k, \varsigma_k, \mathcal{C}_1,...,\mathcal{C}_k)= \sum_{k=1}^K\sum_{i\in \mathcal{C}_k}||x_i-\mu_k-\varsigma_k\beta||^2_{L^2}+\lambda\mathcal{J}(\beta)
\]
By discretising the problem, 

\[
\sum_{k=1}^K\sum_{i\in \mathcal{C}_k}||x_i-\mu_k-\varsigma_k\beta||^2_{L^2}\approx \sum_{k=1}^K\sum_{i\in \mathcal{C}_k} (\boldsymbol{a}_i-\boldsymbol c_k-\varsigma_k\boldsymbol b)'D_0(\boldsymbol a_i-\boldsymbol c_k-\varsigma_k\boldsymbol b) + \lambda \boldsymbol b' D_2 \boldsymbol b
\]
The grandient of the discretised $\mathcal{L}_c$ with respect to $\boldsymbol b$ is given by:

\begin{equation*}
\begin{split}
\nabla_{\boldsymbol{b}}\mathcal{L}_c(\boldsymbol{b}|\boldsymbol c, \boldsymbol\varsigma,\mathcal{C}_1,...,\mathcal{C}_k)=&D_0\cdot\bigg(\sum_{k=1}^K\sum_{i\in \mathcal{C}_k} \varsigma_k(\boldsymbol a_i-\boldsymbol c_k-\varsigma_k\boldsymbol b)\bigg) + \lambda D_2 \cdot \boldsymbol{b}\\
   =&\bigg( \sum_{k=1}^K n_k \varsigma_k^2 D_0+ \lambda D_2 \bigg)\boldsymbol{b}-\bigg( \sum_{k=1}^K n_k \varsigma_k(\bar{\boldsymbol a}_k - \boldsymbol{c}_k)D_0\bigg)
\end{split}
\end{equation*}
Imposing the gradient equal to $0$, we obtain Equation \eqref{eq:update_beta}.\\

\qed

\bibliography{kmodbib_with_doi}

@article{gauglitz10,
  author  = {Gauglitz, G.},
  title   = {Direct optical detection in bioanalysis: an update},
  journal = {Analytical and Bioanalytical Chemistry},
  year    = {2010},
  volume  = {398},
  pages   = {2363--2372},
  doi     = {10.1007/s00216-010-3904-4}
}

@book{horvath12,
  author    = {Horv{\'a}th, L. and Kokoszka, P.},
  title     = {Inference for functional data with applications},
  year      = {2012},
  publisher = {Springer}
}

@book{kolmogorov1961,
  author    = {Kolmogorov, A. N. and Fomin, S. V.},
  title     = {Elements of the Theory of Functions and Functional Analysis},
  year      = {1961},
  publisher = {Graylock Press}
}

@inproceedings{psychometry,
  author    = {Li, Yifan and Li, Binghua and Ding, Jinhong and Feng, Yuan and Ma, Ming and Han, Zerui and Xu, Yehan and Xia, Likun},
  title     = {A Novel Framework for Forecasting Mental Stress Levels Based on Physiological Signals},
  booktitle = {Neural Information Processing. ICONIP 2023},
  series    = {Communications in Computer and Information Science},
  volume    = {1963},
  pages     = {287--297},
  year      = {2024},
  publisher = {Springer, Singapore},
  doi       = {10.1007/978-981-99-8138-0_23}
}

@article{neuroscience,
  author  = {Movahed, R. A. and Pirzad, G. J. and Shahyad, S. and Meftahi, G. H.},
  title   = {A major depressive disorder classification framework based on {EEG} signals using statistical, spectral, wavelet, functional connectivity, and nonlinear analysis},
  journal = {Journal of Neuroscience Methods},
  year    = {2021},
  volume  = {358},
  pages   = {109209},
  doi     = {10.1016/j.jneumeth.2021.109209}
}

@book{ramsaysilvermanbook,
  author    = {Ramsay, J. O. and Silverman, B. W.},
  title     = {Functional Data Analysis},
  year      = {2005},
  publisher = {Springer, New York}
}

@article{medical,
  author  = {Ruan, H. and Dai, X. and Chen, S. and Qiu, X.},
  title   = {Arrhythmia Classification and Diagnosis Based on {ECG} Signal: A Multi-Domain Collaborative Analysis and Decision Approach},
  journal = {Electronics},
  year    = {2022},
  volume  = {11},
  number  = {19},
  pages   = {3251},
  doi     = {10.3390/electronics11193251}
}

@article{engineering,
  author  = {Zhou, R. and Serban, N. and Gebraeel, N.},
  title   = {Degradation Modeling Applied to Residual Lifetime Prediction Using Functional Data Analysis},
  journal = {Annals of Applied Statistics},
  year    = {2011},
  volume  = {5},
  number  = {2B},
  pages   = {1586--1610},
  doi     = {10.1214/10-AOAS448}
}

@article{zhang2022reviewclusteringmethodsfunctional,
  author  = {Zhang, M. and Parnell, A.},
  title   = {Review of Clustering Methods for Functional Data},
  journal = {ACM Transactions on Knowledge Discovery from Data},
  year    = {2023},
  volume  = {17},
  number  = {7},
  pages   = {1--34},
  doi     = {10.1145/3581789}
}

@article{JaquePreda,
  author  = {Jacques, J. and Preda, C.},
  year    = {2014},
  pages   = {231--255},
  title   = {Functional Data Clustering: A Survey},
  volume  = {8},
  journal = {Advances in Data Analysis and Classification},
  doi     = {10.1007/s11634-013-0158-y}
}

@Inbook{Bock2007,
  author    = {Bock, H. H.},
  editor    = {Brito, P. and Cucumel, G. and Bertrand, P. and de Carvalho, F.},
  title     = {Clustering Methods: A History of k-Means Algorithms},
  booktitle = {Selected Contributions in Data Analysis and Classification},
  year      = {2007},
  publisher = {Springer Berlin Heidelberg},
  address   = {Berlin, Heidelberg},
  pages     = {161--172},
  doi       = {10.1007/978-3-540-73560-1_15}
}

@inbook{Schubert_2019,
  title     = {Faster k-Medoids Clustering: Improving the PAM, CLARA, and CLARANS Algorithms},
  booktitle = {Similarity Search and Applications},
  publisher = {Springer International Publishing},
  author    = {Schubert, E. and Rousseeuw, P. J.},
  year      = {2019},
  pages     = {171--187},
  doi       = {10.1007/978-3-030-32047-8_16}
}

@article{Wu18102022,
  author    = {Ruhao, W. and Bo, W. and Aiping, X.},
  title     = {Functional data clustering using principal curve methods},
  journal   = {Communications in Statistics - Theory and Methods},
  volume    = {51},
  number    = {20},
  pages     = {7264--7283},
  year      = {2022},
  publisher = {Taylor \& Francis},
  doi       = {10.1080/03610926.2021.1872636}
}

@article{Peng_2008,
  title     = {Distance-based clustering of sparsely observed stochastic processes, with applications to online auctions},
  volume    = {2},
  number    = {3},
  pages     = {1056--1077},
  journal   = {The Annals of Applied Statistics},
  publisher = {Institute of Mathematical Statistics},
  author    = {Peng, J. and M{\"u}ller, H. G.},
  year      = {2008},
  doi       = {10.1214/08-AOAS172}
}

@article{FRECL,
  author    = {Conde, Susana and Tavakoli, Shahin and Ezer, Daphne},
  journal   = {PLOS ONE},
  publisher = {Public Library of Science},
  title     = {Functional regression clustering with multiple functional gene expressions},
  year      = {2024},
  number     = {11},
  volume    = {19},
  pages     = {e0310991},
  doi       = {10.1371/journal.pone.0310991}
}

@article{SSscore,
  author  = {Rousseeuw, P.},
  year    = {1987},
  pages   = {53--65},
  title   = {Silhouettes: A Graphical Aid to the Interpretation and Validation of Cluster Analysis},
  volume  = {20},
  journal = {Journal of Computational and Applied Mathematics},
  doi     = {10.1016/0377-0427(87)90125-7}
}

@article{Patane2026,
  author    = {Patanè, G. and Nicolussi, F. and Krauth, A. and 
               Gauglitz, G. and Colosimo, B. M. and Dede', L. and 
               Menafoglio, A.},
  title     = {Functional-Ordinal Canonical Correlation Analysis with 
               Application to Data from Optical Sensors},
  journal   = {Technometrics},
  volume    = {68},
  number    = {2},
  pages     = {384--398},
  year      = {2026},
  publisher = {Taylor \& Francis},
  doi       = {10.1080/00401706.2025.2584479}
}

@article{Oti2021,
  author  = {Oti, E. U. and Olusola, M. O. and Eze, F. C. and Enogwe, S. U.},
  title   = {Comprehensive Review of {K}-Means Clustering Algorithms},
  journal = {International Journal of Advances in Scientific Research and Engineering},
  year    = {2021},
  volume  = {7},
  number  = {8},
  pages   = {64--69},
  doi     = {10.31695/IJASRE.2021.34050}
}

@incollection{Kaufman1990,
  author    = {Kaufman, L. and Rousseeuw, P. J.},
  title     = {Partitioning Around Medoids (Program {PAM})},
  booktitle = {Finding Groups in Data: An Introduction to Cluster Analysis},
  publisher = {John Wiley \& Sons},
  year      = {1990},
  pages     = {68--125},
  doi       = {10.1002/9780470316801.ch2}
}

\end{document}